\documentclass[11pt,a4paper]{article}
\usepackage[hyperref]{acl2021}
\usepackage{times}
\usepackage{latexsym}
\usepackage{lipsum}

\usepackage{microtype}
\usepackage{url}
\usepackage{subfigure}
\usepackage{multirow}
\usepackage{enumitem}
\usepackage{stmaryrd}
\usepackage{array}
\usepackage{threeparttable}
\usepackage{blindtext}
\usepackage{amssymb}

\usepackage[ruled,vlined,linesnumbered,longend]{algorithm2e}
\usepackage{svg}
\usepackage{placeins}

\usepackage{soul}
\usepackage[utf8]{inputenc}
\usepackage{graphicx}
\usepackage{amsmath}
\usepackage{booktabs}
\newcommand{\hide}[1]{}

\newcommand{\hero}{H{\small IF}}
\newcommand{\kat}{K{\small AT}}
\newcommand{\cfc}{C{\small FC}}

\newcommand{\shero}{H{\small IF} }
\newcommand{\skat}{K{\small AT} }
\newcommand{\scfc}{C{\small FC} }

\usepackage{amsmath,amsfonts,bm}

\def\eqref#1{equation~\ref{#1}}

\def\1{\bm{1}}

\def\rva{{\mathbf{a}}}

\def\rvk{{\mathbf{k}}}

\def\rvq{{\mathbf{q}}}

\def\rvv{{\mathbf{v}}}

\def\rmA{{\mathbf{A}}}

\def\rmK{{\mathbf{K}}}

\def\rmQ{{\mathbf{Q}}}

\def\rmV{{\mathbf{V}}}
\def\rmW{{\mathbf{W}}}

\DeclareMathAlphabet{\mathsfit}{\encodingdefault}{\sfdefault}{m}{sl}
\SetMathAlphabet{\mathsfit}{bold}{\encodingdefault}{\sfdefault}{bx}{n}

\def\gA{{\mathcal{A}}}

\def\sR{{\mathbb{R}}}

\aclfinalcopy 
\def\aclpaperid{4111} 

\title{

Interpretable and Low-Resource Entity Matching \\
via Decoupling Feature Learning from Decision Making

}

\newcommand*{\email}[1]{\texttt{#1}}

\author{
Zijun Yao$^{1,2}$ \quad Chengjiang Li$^{1,2}$ \quad Tiansi Dong$^{3}$ \quad Xin Lv$^{1,2}$ \quad Jifan Yu$^{1,2}$ \\
{\bf Lei Hou}$^{1,2}$\thanks{\quad Corresponding to L.Hou (houlei@tsinghua.edu.cn)} \qquad {\bf Juanzi Li}$^{1,2}$ \qquad {\bf Yichi Zhang}$^{4}$ \qquad {\bf Zelin Dai}$^{4}$ \\
$^1$Department of Computer Science and Technology, BNRist; \\
$^2$KIRC, Institute for Artificial Intelligence \\
Tsinghua University, Beijing 100084, China \\
$^3$B-IT, University of Bonn, Germany \\
$^4$Alibaba Group, Hangzhou, China \\
\email{\{yaozj20@mails., houlei@\}tsinghua.edu.cn} \\
\email{dongt@bit.uni-bonn.de}\\
}

\date{}

\begin{document}
\maketitle

\begin{abstract}
    Entity Matching (EM) aims at recognizing entity records that denote the same real-world object. 
    Neural EM models  
    learn vector representation of entity descriptions and match entities end-to-end.
    Though robust, 
    these methods require many 
    annotated resources for training, and lack of interpretability.
    In this paper, we propose a novel EM framework that consists of Heterogeneous Information Fusion (\hero) and Key Attribute Tree (\kat) Induction to decouple feature representation from matching decision. 
    Using self-supervised learning and mask mechanism in pre-trained language modeling, \shero learns the embeddings of noisy  
    attribute values by 
    inter-attribute attention with unlabeled data. 
Using a set of comparison features and a limited amount of annotated data, \skat Induction learns an efficient decision tree that can be interpreted by generating entity matching rules whose structure is advocated by domain experts. 
    Experiments on 6 public datasets and 3 industrial datasets show that our method is highly efficient and outperforms 
    SOTA EM models in most cases. 
    Our codes and datasets can be obtained from \url{https://github.com/THU-KEG/HIF-KAT}.
\end{abstract}

\section{Introduction}
\begin{figure}[t]
		\centering
		\includegraphics[width = 1.02\linewidth]{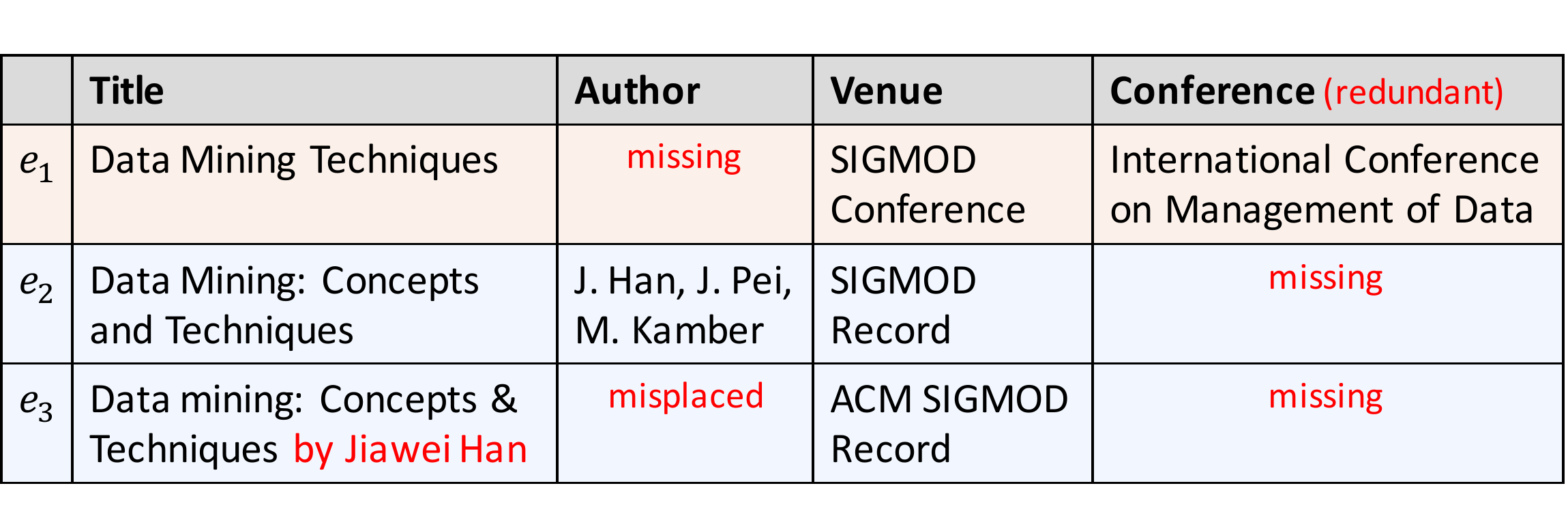}
        \caption{Published papers as entity records.
        }\label{fig:def}
\end{figure}

Entity Matching (EM) aims at identifying whether two records from different sources refer to the same real-world entity. This is a fundamental research task in 
knowledge graph integration~\citep{dong2014knowledge,daniel2020eager,christophides2015entity,christendata} and text mining~\citep{zhao2014group}. 
In real applications, it is not easy 
to decide whether two records with ad hoc linguistic descriptions refer to the same entity. 
In Figure~\ref{fig:def}, $e_2$ and $e_3$ refer to the same publication, while $e_1$ refers to a different one. \textit{Venue}s of $e_2$ and $e_3$ have different expressions; \textit{Author}s of $e_3$ is misplaced in its \textit{Title} field. 

Early works include feature engineering~\citep{wang2011entity} and rule matching~\citep{singh2017generating, fan2009reasoning}.
Recently, 
the robustness of Entity Matching has been improved by deep learning models, 
such as distributed representation based models~\citep{ebraheem2018distributed}, attention based models~\citep{mudgal19deepmatcher,fu2019end,fu2020hierarchical}, and pre-trained language model based models~\citep{li2020deep}.
Nevertheless, these modern neural EM models suffer from two limitations as follows. 

\vspace{-0.1in}
\paragraph{Low-Resource Training.}
Supervised deep learning EM 
relies on large amounts of labeled training data, which is extremely costly in reality.
Attempts have been made 
to leverage external data via transfer learning~\citep{zhao10autoem,thirumuruganathan2018reuse,kasai2019low,loster2021knowledge} and pre-trained language model based methods~\citep{li2020deep}. 
Other attempts have also been made to improve labeling efficiency via active learning~\citep{nafa2020active} and crowdsourcing techniques~\citep{gokhale2014hands,wang2012crowder}.
However, external information may introduce noises, and active learning and crowdsourcing still require additional labeling work.


\vspace{-0.1in}
\paragraph{Lack of Interpretability.}
It is important to know why two entity records are equivalent~\citep{chen2020towards},
however, deep learning EM lacks interpretability. 
Though some neural EM models analyze the model behavior from the perspective of attention~\citep{nie2019deep}, 
attention is not a safe indicator for interpretability~\citep{serrano2019attention}. Deep learning EM also fails to generate interpretable EM rules in the sense that they meet the criteria by domain experts~\citep{fan2009reasoning}.

To address the two limitations, we  propose a novel EM framework to decouple feature representation from matching decision. Our framework consists of Heterogeneous Information Fusion (\hero) and Key Attribute Tree (\kat) Matching Decision for low-resource settings. \shero is robust for \textit{feature representation} from  noisy inputs, and \skat carries out interpretable decisions for entity matching. 


In particular, \shero learns from unlabeled data a mapping function, which converts each noisy attribute value of entity into a vector representation. This is carried out by a novel self-supervised attention  training schema to leverage the redundancy within attribute values and propagate information across attributes. 

\skat Matching Decision learns KAT using decision tree classification. 
After training, \skat carries out entity matching as a task of the classification tree.
For each entity pair, it first computes multiple similarity scores for each attribute using a family of metrics and concatenates them into a comparison feature vector. 
This classification tree can be directly interpreted as EM rules that share a similar structure with EM rules derived by domain experts. 


Our EM method achieves at least SOTA performance on 9 datasets (3 structured datasets, 3 dirty datasets, and 3 industrial datasets) under various extremely low-resource settings.
Moreover, when the number of labeled training data decreases from 60\% to 10\%, our method achieves almost the same performance. In contrast, other methods' performances decrease greatly. 

The rest of the paper is structured as follows. Section~\ref{sec:definition} 
defines the EM task; Section~\ref{sec:method} presents \shero and \kat-Induction in details; Section~\ref{sec:experiments} reports a series of comparative experiments that show the robustness and the interpretability our methods in low-resource settings; Section~\ref{sec:relate} lists some related works; Section~\ref{sec:conclusion} concludes the paper. 





\section{Task Definitions}\label{sec:definition}

\paragraph{Entity Matching.}
Let $T_1$ and $T_2$ be two collections of entity records with $m$ aligned attributes $\{\gA_1, \cdots \gA_m\}$. 
We denote the $i^{th}$ attribute values of entity record $e$ as $e[\gA_i]$.
Entity matching aims to determine whether 
$e_1$ and $e_2$ refer to the same real-world object 
or not. 
Formally, entity matching is viewed as a binary classification function $T_1 \times T_2\rightarrow\{True, False\}$  that takes $(e_1, e_2) \in T_1 \times T_2$ as input, and outputs $True$ ($False$), if $e_1$ and $e_2$ are matched (not matched). 


Current neural EM approaches simultaneously embed entities in low-dimensional vector spaces and obtain entity matching by computations on their vector representations. 
Supervised deep learning EM  relies on large amounts of labeled training data, which is time-consuming and needs costly manual efforts. Large unlabelled data also contain entity feature information useful for EM, yet has not been fully exploited by the existing neural EM methods. 
In this paper, we  aim at decoupling feature representation from matching decision. Our novel EM model consists of two sub-tasks: learning feature representation from unlabeled data and EM decision making.

\paragraph{Feature Representation from Noisy Inputs.}

Entity records are gathered from different sources with three typical noises in attribute values:
{\em misplacing}, {\em missing}, or {\em synonym}.
{\em Misplacing} means that attribute value of $\gA_i$ drifts to $\gA_j (i \neq j)$; 
{\em missing} means that attribute values are empty; 
{\em synonym} means that attribute values with the same meaning have different literal forms. 
Our first task is to fusion noisy heterogeneous information in a self-supervised manner with unlabelled data.




\paragraph{Interpretable EM.}

Domain experts have some valuable specifications on EM rules as follow: (1) an EM rule is an {\em if-then} rule of feature comparison;  (2) it only selects a part of key attributes from all entity attributes for decision making; (3) feature comparison is limited to a number of similarity constraints, such as $=$, $\approx$~\citep{fan2009reasoning,singh2017generating}. 
Our second task is to realize an interpretable EM decision process by comparing feature representation per attribute by utilizing a fixed number of quantitative similarity metrics and then training a decision tree using a limited amount of labeled data.
Our interpretable EM decision making will ease the collaboration with domain experts. 

\begin{figure*}[!ht]
	\centering
	\includegraphics[width = 0.94\linewidth]{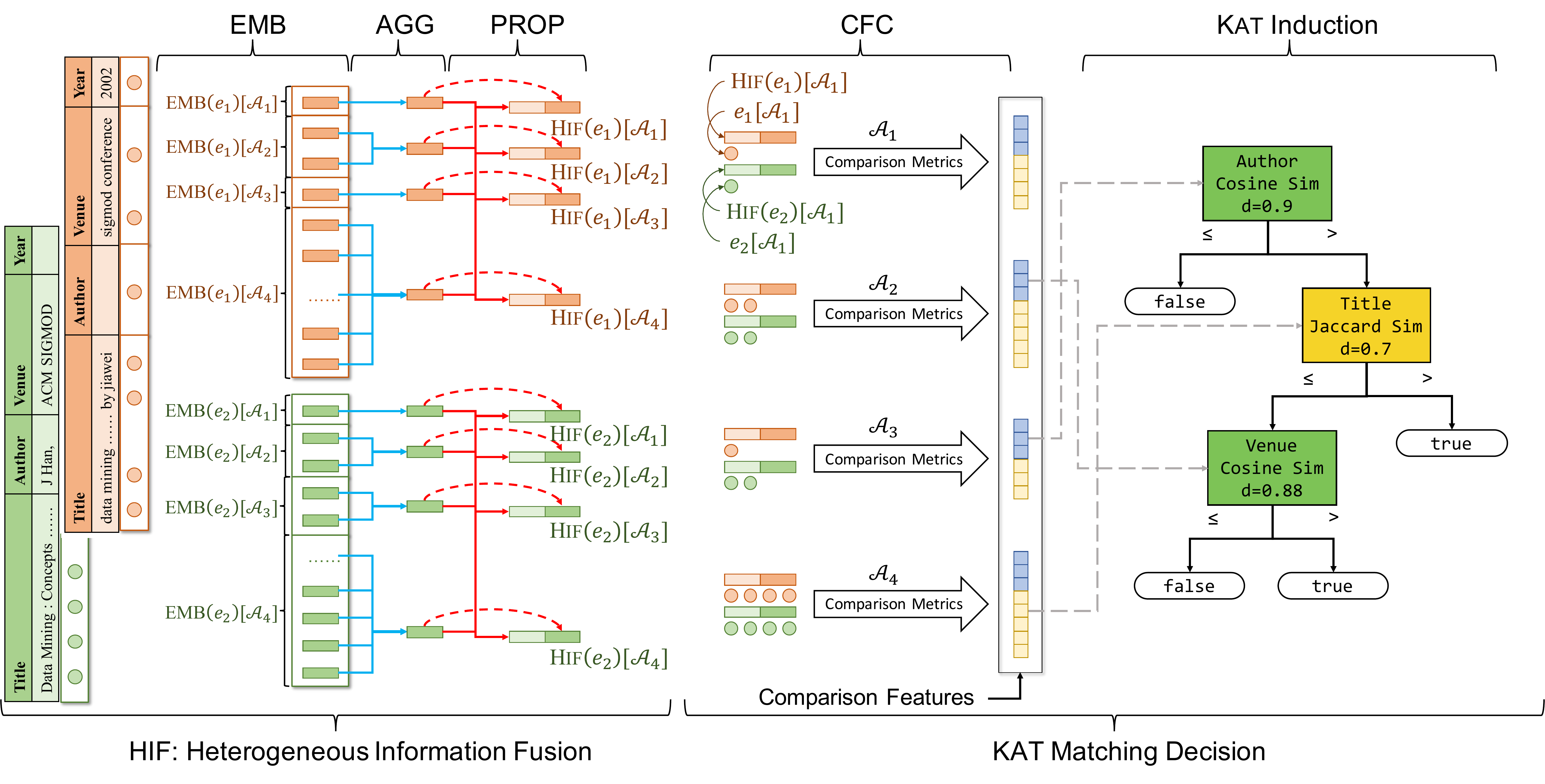}
    \vspace{-0.1in}
    \caption{
    The decoupled EM model comprising 
    the heterogeneous information fusion module
    and the matching decision making module.
    We use circles  and rectangles to denote words and vectors, respectively.
    {\color{cyan}Cyan} lines with arrow indicate word information aggregation via intra-attribute attention.
    {\color{red}Red} lines with arrow show attribute information propagation. 
    In the comparison features vector, \textcolor[RGB]{42,75,160}{blue} squares are similarity scores by comparing on $\text{\hero}(e_1)[\gA_i], \text{\hero}(e_2)[\gA_i]$ and \textcolor[RGB]{183,133,0}{yellow} squares are similarity scores by comparing on $e_1[\gA_i], e_2[\gA_i]$ directly. EMB, AGG, PROP, \cfc, and \kat-Induction are calculation components specified in Section~\ref{sec:method}.
    }\label{fig:model}
\end{figure*}


\section{Methodology}\label{sec:method}

In this section, we introduce (1) a neural model, Heterogeneous Information Fusion (\hero), for the task of feature representation, and (2) a decision tree, Key Attribute Tree (\kat), for the task of interpretable EM.  
Figure~\ref{fig:model} illustrates 
the overall workflow 
of our method.
The following subsections dive into details of the two tasks and propose a novel training scheme for low resource settings by exploiting unlabelled entity records.


\subsection{\shero for Entity Attribute Embedding}

$\text{\hero}: T\rightarrow\sR^{m \times d}$ is a function that maps entity records into vector representations. 
An attribute value $e[\gA_i]$ of a record $e$ is mapped to a $d$ dimensional vector, written as $\text{\hero}(e)[\gA_i] \in \sR^{d}$.
\shero treats attribute values as strings of words and performs word embedding (EMB), word information aggregation (AGG), and attribute information propagation (PROP) successively.


\paragraph{Word Embedding (EMB).}
Word embedding is a pre-train language model that contains features learned from a large corpus. 
We convert numerical and encoded attribute values into strings of digits or alphabets.
For Chinese attribute values, we do word-segmentation using {\em pkuseg} \citep{pkuseg}. 
Then, we mark the beginning and the end of an attribute value with two special tokens, namely $\langle\text{BEG}\rangle$ and $\langle\text{END}\rangle$.
Finally, we pad each attribute value with $\langle\text{PAD}\rangle$ so that they are represented in the same length $l$. The representation after padding is illustrated as below:
$$
\underbrace{\left(\langle\text{BEG}\rangle, w_1, w_2, \cdots \langle\text{END}\rangle, \langle\text{PAD}\rangle, \cdots, \langle\text{PAD}\rangle\right)}_{\text{length}\ =\ l}
$$

Let $W$ be the set of words, each  word $w\in W$ is mapped 
into a vector, and each attribute value is mapped into a matrix.
Formally, $\text{EMB}: W^N\rightarrow\sR^{N \times d_e}$ maps $N$ words into an $N \times d_e$ matrix by executing a look-up-table operation. $N$ is the dictionary size. In particular, we have $\text{EMB}(e)[\gA_i] \in \sR^{l \times d_e}$, in which $d_e$ is the dimension of word embedding vectors. 
It is worth noting that $\langle\text{PAD}\rangle$ is embedded to zero vector to ensure that it does not interfere with other non-padding words in the following step. 

\paragraph{Word Information Aggregation (AGG).}

Summing up the $l$ word embeddings as the embedding of an attribute value
will neglect the importance weight 
among the $l$ words. We leverage a more flexible framework, which aggregates word information by weighted pooling.
The weighting coefficients $\alpha_i$ for different words are extracted by multiplying its embedding vector with a learnable, and attribute-specific vector $\rva_i \in \sR^{d_e \times 1}$. Subscript $i$ implies that $\alpha_i$ and $\rva_i$ are associated with the $i^{th}$ attribute $\gA_i$.
The weighting coefficients are normalized by {\em Softmax} function among words.
Finally, we enable a non-linear transformation (e.g., ReLU) during information aggregation with 
parameters $\rmW_{ai} \in \sR^{d_e \times d_a}$. Formally, $\text{AGG}$ maps each attribute value of entity record $e$ into a $d_a$ dimensional vector $\text{AGG}(\text{EMB}(e)[\gA_i]) \in \sR^{d_a}$ as below:
\begin{equation*}\small 
    \text{AGG}(\text{EMB}(e)[\gA_i]) = \text{ReLU}\left(\alpha_i\ \text{EMB}(e)[\gA_i]\ \rmW_{ai}\right)
\end{equation*}
\begin{equation*}
    \alpha_i = \text{Softmax}(\text{EMB}(e)[\gA_i]\ a_i)^\top \in \sR^{1 \times l}
\end{equation*}

\paragraph{Attribute Information Propagation (PROP).}
The mechanism of attribute information propagation is the key component for noise reduction and representation unification. 
This mechanism is inspired by the observation that missing attribute values often 
appear in other attributes (e.g., \textit{Venue} and \textit{Conference} in Figure~\ref{fig:def}, \citet{mudgal19deepmatcher} also reported the misplacing issue).

We use ``Scaled Dot-Product Attention''~\citep{vaswani2017attention} to propagate information among different attribute values. 
We use parameters $\rmQ, \rmK, \rmV_i$ to convert $\text{AGG}(\text{EMB}(e)[\gA_i])$ into query, key, and value vectors, respectively (Notice that only $\rmV_i$ is attribute-specific).
$\rmA \in \sR^{m \times m}$ is the attention matrix.
$\rmA_{ij}$ denotes the attention coefficients from the $i^{th}$ attribute to the $j^{th}$ attribute:
\begin{align*}
    \rmA_{ij} &= \text{Softmax}
    \left(\frac{\rvq_i \cdot \rvk_j}{\sqrt{m}}\right)\\
    \rvq_i &= \text{AGG}(\text{EMB}(e)[\gA_i])\ \rmQ \\
    \rvk_j &= \text{AGG}(\text{EMB}(e)[\gA_i])\ \rmK \\
    \rvv_i &= \text{AGG}(\text{EMB}(e)[\gA_i])\ \rmV_i
\end{align*}
Record notation $e$ is omitted in vectors $\rvq, \rvk, \rvv$ for brevity.
To keep the identity information, each attribute value after attribute information propagation is represented by the concatenation of the context and the value vector:
\begin{equation*}\label{equ:prop}\small
    \text{PROP}(\text{AGG}(e))[\gA_i] = \text{ReLU}\left(\rvv_i \left\Vert \sum_{j \neq i}\rmA_{ij} \rvv_j \right.\right)
\end{equation*}
\noindent \shero  outputs with Multiple Layer Perceptron (MLP). The whole process can be summarized as follows:
\begin{equation*}
    \text{\hero}(e) = \text{MLP} \circ \text{PROP} \circ \text{AGG} \circ \text{EMB}(e) \in \sR^{m \times d}
\end{equation*}
After \hero, each attribute $\gA_i$ of an entity record $e$ has a feature embedding 
$\text{\hero}(e)[\gA_i]$. 
 
\subsection{\skat for Matching Decision}


\skat Matching Decision consists of two steps: comparison feature computation (\cfc) and decision making with \kat. \scfc computes similarity score for each paired attribute features by utilizing a family of well-selected metrics, and concatenate these similarity scores into a vector (comparison feature). 
\skat takes comparison feature as inputs, and perform entity matching with a decision tree.



\paragraph{Comparison Feature Computing (CFC).}
Given a record pair $(e_1, e_2)$, \scfc implements a function 
that maps $(e_1, e_2)$ to a vector of similarity scores $\text{\cfc}(e_1, e_2)$. 
The similarity score $\text{\cfc}(e_1, e_2)$ is a concatenation of a similarity vector between paired attribute values (i.e., $e_1[\gA_i], e_2[\gA_i]$) and a similarity vector between their vector embeddings (i.e., $\text{\hero}(e_1)[\gA_i], \text{\hero}(e_2)[\gA_i]$).

To compare paired attribute values, we follow \citet{conda16magellan} and classify attribute values into 6 categories, according to the type and the length, each with a set of comparison metrics for similarity measurement, such as 
Jaccard similarity, Levenshtein similarity, Monge-Elkan similarity, etc. More details are presented in Table~\ref{tab:app:metrics}.

For attribute value embeddings, we choose three 
metrics: 
the cosine similarity, the $L_2$ distance, and the Pearson coefficiency. 
In this way, we 
convert entity record pair into similarity score vector of attributes. Each dimension indicates the similarity degree of one attribute 
from a certain perspective.

\begin{table*}[t]
    \centering
    \small
    \scalebox{1.0}{
    \begin{tabular}{l|cc}
    \toprule
    Attribute Type  & Comparison Metrics \\
    \midrule
    \multirow{1}{*}{boolean} 
        & Exact matching distance \\
    \midrule
    \multirow{2}{*}{number} 
        & Exact matching distance, Absolute distance, \\
        & Levenshtein distance, Levenshtein similarity \\
    \midrule
    \multirow{3}{*}{string of length $1$} 
        & Levenshtein distance, Levenshtein similarity,\\
        & Jaro similarity, Jaro Winkler similarity, \\
        & Exact matching distance, Jaccard similarity with QGram tokenizer,\\
    \midrule
    \multirow{4}{*}{string of length $[2, 5]$} 
        & Jaccard similarity with QGram tokenizer, Jaccard similarity with delimiter tokenizer,\\
        & Levenshtein distance, Levenshtein similarity\\
        & Cosine similarity with delimiter tokenizer,\\
        & Monge Elkan similarity, Smith Waterman similarity,\\
    \midrule
    \multirow{3}{*}{string of length $[6, 10]$} 
        & Jaccard similarity with QGram tokenizer, Cosine similarity with delimiter tokenizer,\\
        & Levenshtein distance, Levenshtein similarity, \\
        & Monge Elkan similarity \\
    \midrule
    \multirow{1}{*}{string of length $[10, \infty]$} 
        & Jaccard similarity with QGram tokenizer, Cosine similarity with delimiter tokenizer\\
    \bottomrule
    \end{tabular}
    }
    \caption{
    The attributes are classified into 6 categories according to their type and their string lengths.
    For different types of attributes, we use different comparison metrics.
    }
    \label{tab:app:metrics}
\end{table*}



\paragraph{\skat Induction.}

In the matching decision, we take $\text{\cfc}(e_1, e_2)$ as input, and output binary classification results.
We propose Key Attribute Tree, a decision tree, to make the matching decision based on \textit{key attribute} heuristic, in the sense that some attributes are more important than others for EM. 
For example, we can decide whether two records of research articles are the same by only checking their \textit{Title} and \textit{Venue} without examining their \textit{Conference}.
Focusing only on key attributes not only saves computations, but also introduces interpretability that has two-folded meanings: 
(1) each dimension of $\text{\cfc}(e_1, e_2)$  is a candidate feature matching which can be interpreted as a component of an EM rule; 
(2) the decision tree learned by \skat 
can be converted into EM rules that follow the same heuristics as the EM rules made by domain experts~\citep{fan2009reasoning}.  

\subsection{Model Training}

HIF and KAT Induction are trained separately.

\paragraph{\shero Training.}

We design a self-supervised training method for \shero to learn from unlabeled data.
Our strategy is 
to let the \shero model 
predict manually masked attribute values. 
We first represent attribute values, as strings of words, by Weighted Bag Of Words (WBOW) vectors, 
whose dimensions 
represent word frequencies. 
Then, we manually corrupt a small portion of entity records in $T_1 \cup T_2$ by randomly replacing (mask) their attribute values with an empty string, which forms a new table $T'$.
\shero takes $T'$ as input and uses another MLP to predict the WBOW of masked attribute values.
\shero is trained by minimizing the Cross-Entropy 
between the prediction and the ground-truth WBOW:
\begin{equation*}
    \min_{\text{H\tiny IF}} \text{CrossEntropy}\left(\text{MLP}(\text{\hero}(T')), \text{WBOW}\right)
\end{equation*}

\paragraph{\skat Induction Training.}

\skat is trained with a normal decision tree algorithm. 
We constrain its 
depth, in part to maintain the interpretability of transformed EM rules. 
We use xgboost~\citep{chen2016xgboost} and ID3 algorithm~\citep{quinlan1986induction} in the experiments. To preserve interpretability, the booster number of xgboost is set to 1, which means it only learns one decision tree.
For $(e_1, e_2, True) \in D$, \skat takes $\text{\cfc}(e_1, e_2)$ as input, and $True$ as the target classification output.

\section{Experiments}\label{sec:experiments}



\subsection{Experimental Setup}

\begin{table}[t]
\centering
\scalebox{0.81}{
\setlength{\tabcolsep}{3.8pt} 
\begin{tabular}{cccccccc}
	\toprule
	Type & Dataset & \#Attr. & \#Rec. & \#Pos. & \#Neg. & Rate \\
	\midrule
	\multirow{3}{*}{{Structured}}
	&I-A$_1$        & 8 & 2,908 & 132 & 407 & 10\%\\
	&D-A$_1$             & 4 & 4,739 & 2,220 & 10,143 & 1\%\\
	&D-S$_1$         & 4 & 13,270 & 5,347 & 23,360 & 1\% \\
	\midrule
	\multirow{3}{*}{{Dirty} }
	&I-A$_2$        & 8 & 2,908 & 132 & 407 & 10\%\\
	&D-A$_2$             & 4 & 4,739 & 2,220 & 10,143 & 1\%\\
	&D-S$_2$         & 4 & 13,270 & 5,347 & 23,360 & 1\%\\
	\midrule
	\multirow{3}{*}{{Real}} 
	& Phone & 36 & 940 & 1,099 & 2,241 & 10\%\\
	& Skirt & 20 & 9,708 & 6,371 & 18,202 & 1\%\\
	& Toner & 13 & 7,065 & 4,551 & 13,481 &1\%\\
	\bottomrule
\end{tabular}
}
\caption{Statistics of the datasets. \#Attr. is the number of attributes, \#Rec. is the number of entity records, and \#Pos. (\#Neg.) is the number of labeled positive (negative) pairs. I-A indicates matching between iTunes-Amazon. D-A indicates matching between DBLP-ACM. D-S indicates matching between DBLP-Google Scholar. We use subscripts 1, 2 to distinguish between \textit{Structured} and \textit{Dirty} data.}
\label{tab:dataset}
\end{table}
\subsubsection{Datasets}

In order to evaluate our model comprehensively, we collect multi-scaled datasets ranging from English corpus and Chinese corpus, including 
\textit{Structured} datasets, \textit{Dirty} datasets, and \textit{Real} datasets. 
\textit{Structured} and \textit{Dirty} datasets
are benchmark datasets\footnote{\url{http://pages.cs.wisc.edu/~anhai/data1/deepmatcher_data/}\label{dmgithub}} released in \cite{mudgal19deepmatcher}. 
The \textit{Real} datasets are sampled from Taobao---one of the biggest E-commerce platform in China, a portion of which 
are manually labeled to indicate whether they are the same entity or not.
The \textit{real} datasets have notably more attributes than the \textit{structured} or \textit{dirty} datasets.

Statistics of these datasets are listed in Table~\ref{tab:dataset}. 
We focus on setting of \textit{low resource} EM and use Rate\% of labelled data as training set.
The validation set uses the last 20\% labeled pairs, and the rest pairs in the middle are the test set.
This splitting is different from the \textit{sufficient resource} EM~\cite{mudgal19deepmatcher,conda16magellan} where up to 60\% pairs are used in the training set.
For \textit{I-A$_1$}, \textit{I-A$_2$}, and \textit{Phone}, we use 10\% labeled pairs as training data, because some of the baselines will crash, if the training data is too small.

We remove trivial entity pairs from the \textit{Real} datasets, as
\textit{Structured} and \textit{Dirty} datasets have been released. 
For \textit{Real} datasets, we remove matching pairs with large Jaccard similarity (0.32 for Phone, 0.36 for others) and non-matching pairs with small Jaccard similarity (0.3 for Phone, 0.332 for others).


\begin{figure*}[!ht]
	\centering
	\includegraphics[width = 1.\linewidth]{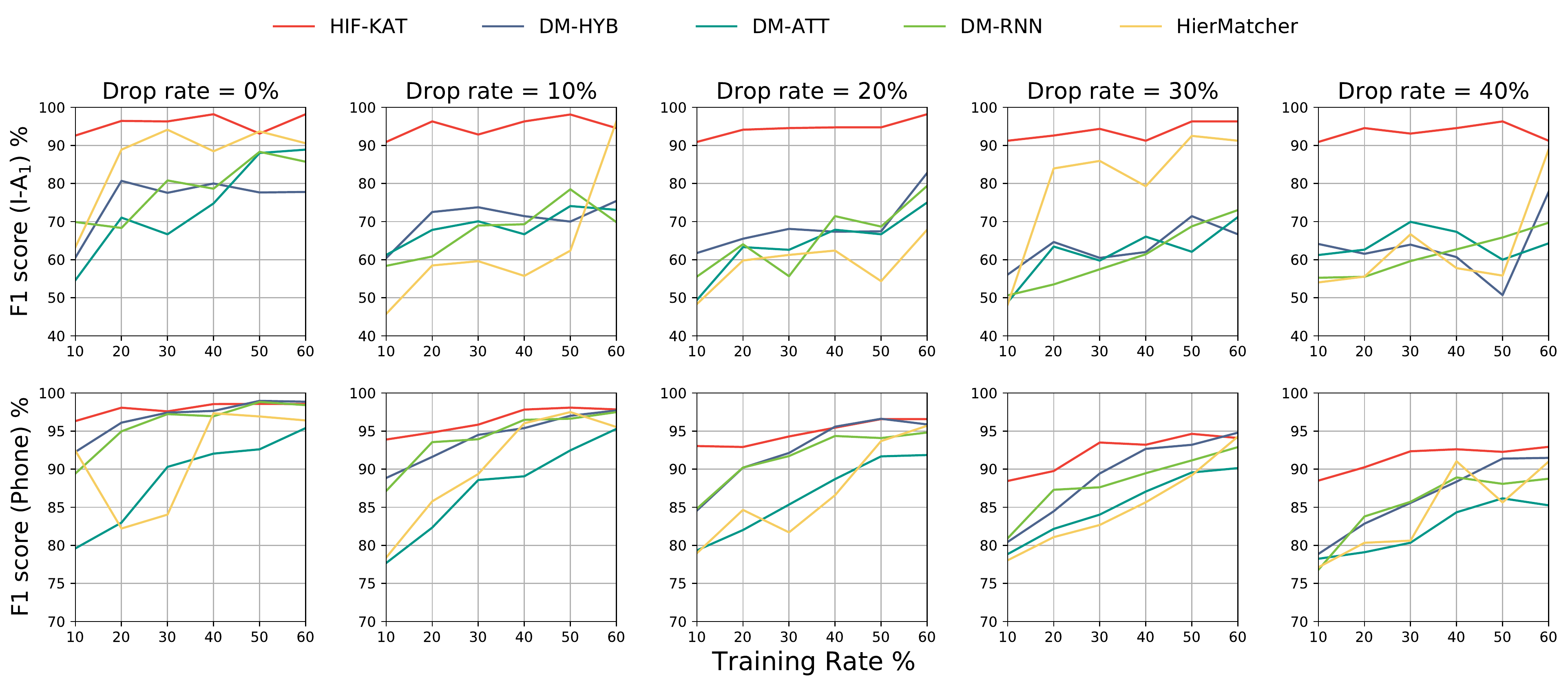}
	\vspace{-0.3in}
    \caption{
    Results for robustness. \hero+\skat refers to \hero+\kat$_{\text{XGB}}$. 
    Each two subgraphs in the same column correspond to the same drop rate (Drop rate is marked on the top of each column).
    Each five subgraphs in the same row correspond to the same dataset. 
    x-axis is the rate of labelled data used in training. y-axis is the F$_1$ score. 
    }\label{fig:robustness}
\end{figure*}

\subsubsection{Baselines}

We implement 3 variants of our methods with different \skat Induction algorithms.
\textbf{\hero+\kat$_{\text{ID3}}$} and \textbf{\hero+\kat$_{\text{XGB}}$}
inducts \skat with ID3 algorithm and xgboost respectively constraining maximum depth to 3.
\textbf{\hero+DT} inducts \skat with ID3 algorithm with no constraints on the tree depth.
We include reproducibility details in Appendix~B.

We compare our methods with three SOTA EM methods, among which two are publicly available end-to-end neural methods, and one is feature engineering based method.
\begin{enumerate} 
    \item \textbf{DeepMatcher} \citep{mudgal19deepmatcher} (DM) is a general deep-learning based EM framework with multiple variants---\textit{RNN} DM{\footnotesize{-RNN}}, \textit{Attention} DM{\footnotesize{-ATT}}, and \textit{Hybrid} DM{\footnotesize{-HYB}}---depending on what building block it chooses to construct\footnote{\url{https://github.com/anhaidgroup/deepmatcher}}.
    \item \textbf{HierMatcher}~\citep{fu2020hierarchical} is also an end-to-end neural EM method that compare entity records at the word level\footnote{\url{https://github.com/cipnlu/EntityMatcher}}.
    \item  \textbf{Magellan}~\citep{conda16magellan} integrates both automatic feature engineering for EM and  classifiers. 
    Decision tree is used as the classifier of Magellan in our experiments.
\end{enumerate}

For ablation analysis, we replace a single component of our model with a new model as follows: \textbf{\hero+LN} replaces \skat with a linear classifier; \textbf{\hero+LR} replaces \skat with a logistic regression classifier; 
\textbf{\hero-ALONE} 
removes comparison metrics of attribute values (yellow segment of comparison features in Figure~\ref{fig:model}). 
We also do ablation analysis for  
\textbf{\hero-ALONE} as follows: \textbf{\hero-WBOW} replaces outputs of \shero with $d$-dimensional WBOW vectors using PCA. \textbf{\hero-EMB} replaces the outputs of \shero with the mean pooling of word embeddings.



\subsubsection{Evaluation Metrics}

We use F$_1$ score as the evaluation metric.
Experiment results are listed in Table~\ref{tab:weak} and Table~\ref{tab:full}. 
All the reported results are averaged over 10 runs with different random seeds.

\subsection{Experimental Results}




\paragraph{General Results.}


We evaluate the performance of our model against 3 SOTA models under low resource settings, where only 1\% or 10\% of the total amount of labeled pairs are used for training (See Table~\ref{tab:dataset}).
Comparative experiment results on the 9 datasets are listed in Table~\ref{tab:weak}.

Our decoupled framework achieves SOTA EM results on all the nine datasets, and demonstrates significant performance 
on \textit{Dirty} datasets, 
with a boosting of 4.3\%, 14.7\%, and 8.4\% in terms of F$_1$ score on I-A$_2$, D-A$_2$, D-S$_2$, compared to the best performance of baselines 
on their corresponding datasets.
Our methods also outperforms all baselines on \textit{Structured} and two \textit{Real} datasets (the same as Magellan on Toner).
The out-performance on \textit{Real} datasets is marginal because attribute values in \textit{Real} datasets are quite standard, which means that our model does not have many chances to fix noisy attribute values. 
Still, our methods achieve a high F$_1$ score ($\geq 94.9$\%) in \textit{Real} datasets. 
These results indicate out methods are both effective under low resource settings and robust to noisy data.


\begin{table}
\centering
\scalebox{0.72}{
\setlength{\tabcolsep}{1.6pt} 
\begin{tabular}{l c c c c c c c c c }
\toprule
{Methods} & {I-A$_1$} & {D-A$_1$} & {D-S$_1$} & {I-A$_2$} & {D-A$_2$} & {D-S$_2$} & {Phone} & {Skirt} & {Toner} \\
\midrule
{ DM\footnotesize{-RNN} } & 63.6 & 85.4 & 74.8 & 42.3 & 45.7 & 39.0 & 90.0 & 67.6 & 68.6 \\
{ DM\footnotesize{-ATT} } & 55.8 & 82.5 & 79.0 & 46.5 & 45.2 & 57.8 & 80.3 & 54.4 & 48.8 \\
{ DM\footnotesize{-HYB} } & 60.9 & 86.6 & 78.0 & 49.5 & 46.2 & 60.4 & 91.9 & 64.2 & 67.4 \\
{ HierMatcher } & 61.9 & 37.5 & 68.2 & 37.8 & 32.6 & 45.8 & 86.2 & 61.7 & 55.2 \\
{ Magellan } & 92.3 & 93.7 & 85.1 & 50.6 & 65.6 & 71.1 & 93.6 & 96.6 & \textbf{97.2} \\
\midrule
{ \hero+DT } & \textbf{96.0} & 96.4 & 87.5 & \textbf{54.9} & 80.1 & 74.2 & \textbf{94.9} & \textbf{96.7} & \textbf{97.2} \\
{ \hero+\kat$_{\text{ID3}}$ } & 95.8 & \textbf{96.6} & \textbf{88.2} & 51.6 & 79.0 & \textbf{79.5} & 94.5 & \textbf{96.7} & \textbf{97.2} \\
{ \hero+\kat$_{\text{XGB}}$ } & 90.6 & 93.3 & 87.9 & 41.5 & \textbf{80.3} & \textbf{79.5} & 94.4 & 96.2 & \textbf{97.2} \\
\midrule
\midrule
{ \hero+LN } & 77.9 & 21.0 & 54.7 & 41.6 & - & 78.5 & 72.2 & 62.8 & 86.0 \\
{ \hero+LR } & 84.2 & 87.1 & 84.6 & 46.5 & - & 68.1 & 87.5 & 41.7 & 62.0 \\
\midrule
{ \hero-WBOW } & 93.0 & 92.7 & 75.4 & 43.2 & 47.9 & 43.7 & 91.6 & 66.3 & 74.0 \\
{ \hero-EMB } & 91.1 & 90.9 & 76.6 & 30.8 & 53.9 & 46.8 & 89.9 & 65.7 & 79.8 \\
{ \hero-ALONE } & 94.6 & 96.1 & 82.9 & 45.6 & 73.5 & 63.2 & 91.8 & 63.0 & 72.9 \\
\bottomrule
\end{tabular}
}
\caption{F$_1$ score of all methods under low resource setting(\%). Dash (-) indicates classifier fails to converge.}
\label{tab:weak}
\end{table}


\paragraph{Effectiveness to Low Resource Settings
}
We reduce the training rate from 60\% to 10\% to see whether our method is sensitive to the number of labeled record pairs as training resources. Experimental results are shown in Figure~\ref{fig:robustness}. 
\hero+\skat (red line) achieves a stable performance as the number of labeled record pairs decreases, while the F$_1$ score of DeepMatcher and HierMatcher decrease simultaneously.
Besides, our methods continuously outperform DeepMatcher and HierMatcher, ranging from low resource setting to sufficient resource setting.
These results indicate that by exploring unlabelled data, \shero alleviates the reliance on labeled record pairs.

\paragraph{Effectiveness to Noisy Heterogeneous Data.}
We manually aggravate the quality of datasets by randomly dropping $p\%$ of attribute values ($p\%$ ranges from 0\% to 40\%), and see to what degree the feature representations delivered by \shero will affect the EM decision matching. 
From left to right, columns of subgraphs in Figure~\ref{fig:robustness} demonstrates results with increasing dropping rate.
On the I-A$_1$ dataset, the influence of dropping rate 
is marginal to \hero+\skat, whose F$_1$ score  fluctuates around 95\%. In contrast,  F$_1$ scores 
of both DeepMatcher and HierMatcher 
will decrease if 
more attribute values are dropped.
On the Phone dataset, the dropping rate's influence is not severe to \hero+\kat, especially when the training rate is low.
These results show that \shero is efficient in recovering noisy heterogeneous inputs.

\subsection{Case Study for Interpretablity}\label{sec:case}

\begin{figure}[!t]
	\centering
	\includegraphics[width = 0.9\linewidth]{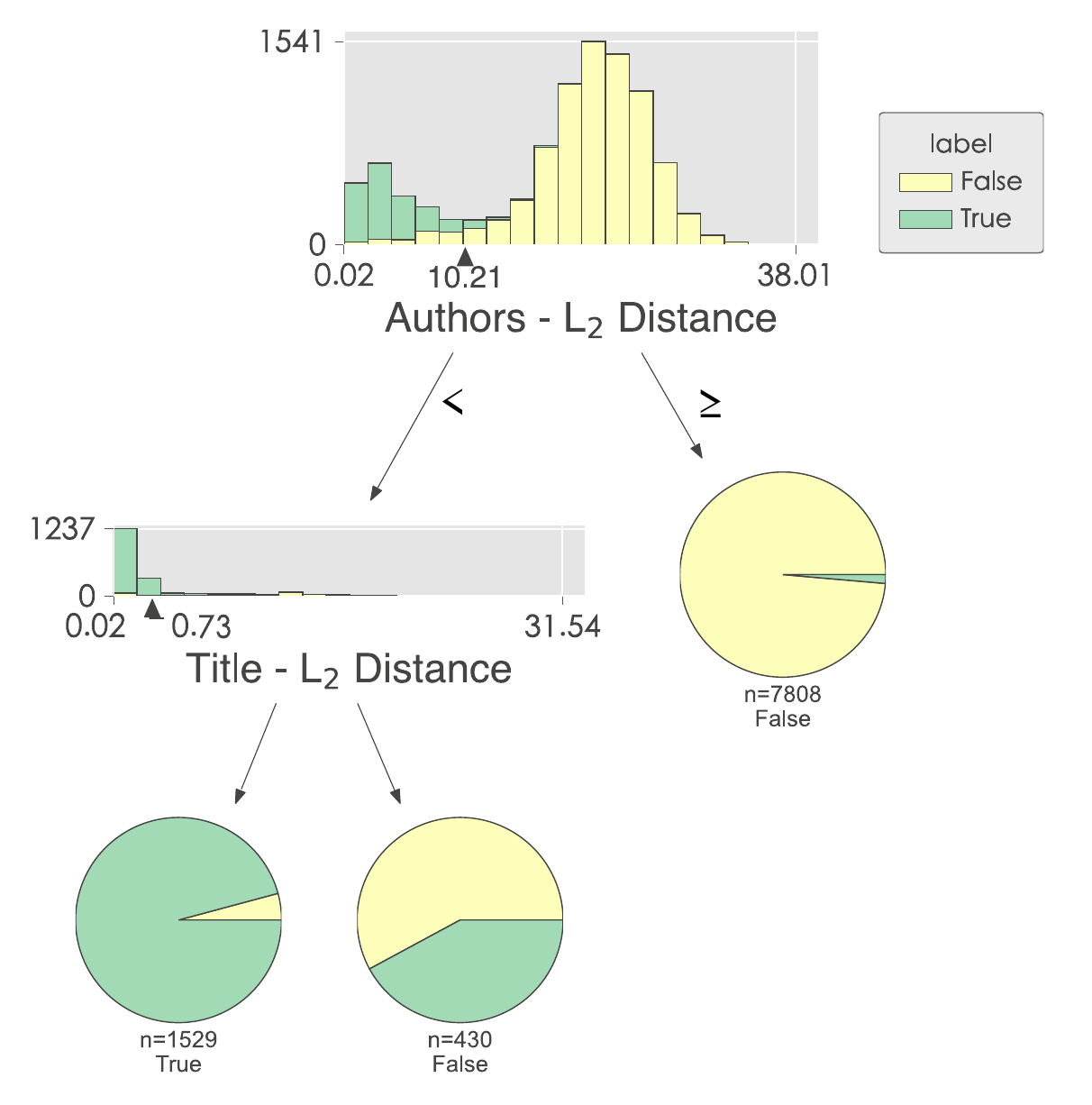}
    \caption{
    The Key Attribute Tree generated by \hero+\kat$_\text{XGB}$ for D-A$_1$ dataset.
    }\label{fig:case}
\end{figure}

The interpretability of our model means that the process of decision making of \skat can be easily transformed into EM rules whose structure is recommended by domain experts. Figure~\ref{fig:case} illustrates a tree decision process of \skat that determines whether two records denote the same  publication in the D-A$_1$ (DBLP and ACM) datasets. 
Each path from the root to a leaf node of the tree structure 
can be converted into an EM rule as follows:
\begin{equation*}\small
\begin{aligned}
    \text{\textsf{{Rule 1: if}}}\ & L_2\left(\text{\hero}(e_1), \text{\hero}(e_2)\right)[\text{Authors}] \geq 10.21 \\
    \text{{\textsf{then}}}\ & e_1, e_2\text{{\textsf{ are not a match;}}}\\
    \text{\textsf{{Rule 2: if}}}\ & L_2\left(\text{\hero}(e_1), \text{\hero}(e_2)\right)[\text{Authors}] < 10.21 \ \\
    & \land L_2\left(\text{\hero}(e_1), \text{\hero}(e_2)\right)[\text{Title}] < 0.73\\
    \text{{\textsf{then}}}\ & e_1, e_2\text{{\textsf{ are a match;}}}\\
    \text{\textsf{{Rule 3: if}}}\ & L_2\left(\text{\hero}(e_1), \text{\hero}(e_2)\right)[\text{Authors}] < 10.21 \\
    &\land  L_2\left(\text{\hero}(e_1), \text{\hero}(e_2)\right)[\text{Title}] \geq 0.73\\
    \text{{\textsf{then}}}\ & e_1, e_2\text{{\textsf{ are not a match}}}\\
\end{aligned}
\end{equation*}
They can be further read as descriptive rules:\\
{\em Rule 1: if two records have different authors, they will be different publications.}\\
{\em Rule 2: if two records have similar authors and similar titles, they will be the same publication.}\\
{\em Rule 3: if two records have similar authors and dissimilar titles, they will not be the same publication.}\\
The soundness of such 
rules can be examined by our experience.

Important features of \skat 
are as follows: (1) \skat is conditioned on attribute comparison; (2) \skat only selects a few key attributes to compare features. 
In our example, there are 4 attributes, {\em Author, Title, Venue} and {\em Conference} in D-A$_1$ dataset, \skat only selects {\em Title} and {\em Author} for EM decision making. 
The transformed rules meet the specifications of manually designed EM rules of domain experts~\citep{fan2009reasoning,singh2017generating}. This kind of interpretability will ease the collaboration with domain experts, and increase the trustworthiness, compared with uninterpretable end-to-end Deep learning EM models.  

\subsection{Discussions}\label{sec:discuss}

\paragraph{Ablation Analysis.}

Experiment results for ablation models are 
listed in Table~\ref{tab:weak}.
On the one hand, \hero+LN and \hero+LR generally outperforms DeepMatcher and HierMatcher on 7 datasets with on-par performance on 2 \textit{Real} datasets. This indicates that \shero and \scfc together extract better comparison features than end-to-end neural methods under low resource settings.
On the other hand, \hero+LN and \hero+LR are weaker than 
the tree induction classifier, suggesting that \skat is more reliable.

Compared with 
\hero-\kat$_\text{ID3}$, Magellan, and \hero-ALONE, \hero-\kat$_\text{ID3}$ achieves the highest performance, indicating that comparison on both 
attribute value embeddings and the original attribute values are 
important.
Compared with 
\hero-ALONE, \hero-WBOW, and \hero-EMB, 
\hero-ALONE outperforms \hero-WBOW and \hero-EMB on the \textit{Dirty} datasets, showing the positive effects of its information reconstruction.

Finally, comparing 
\hero+\skat with \hero+DT, we find that \hero+\skat has 
better performances than \hero+DT on most of the datasets, except for (I-A$_2$ and Phone). 
This shows that non-key attributes may disturb 
decision making.

\paragraph{Efficiency.}
Table~\ref{tab:efficiency} shows the running times of our methods and of the two neural baselines.
Our methods are highly efficient for 
inference, because our methods are highly parallel 
and are memory-saving.
For example, on Phone datasets our methods can inference in a single batch, while HierMatcher can only run in a batch size of 4 with 24GiB RAM.
The training efficiency of our method is comparable with baselines, because when 
the training data is 
small enough, 
baseline models may 
finish one epoch training with only few batches.

\begin{table}
\centering
\scalebox{0.8}{
\setlength{\tabcolsep}{4.pt} 
\begin{tabular}{cccccccc}
		\toprule
		Epoch & I-A$_1$ & D-A$_1$ & D-S$_1$ & Phone & Skirt & Toner \\
		\midrule
		\multirow{1}{*}{DM{-HYB}}
		& 0.98 & 1.0 & 2.3 & 12.7 & 5.1 & 2.5 \\
		\multirow{1}{*}{HierMatcher}
		& 0.47 & 0.3 & 0.7 & 41.7 & 4.0 & 1.4 \\
		\multirow{1}{*}{\hero+\kat$_{\text{ID3}}$}
		& 0.45 & 1.0 & 1.5 & 2.2 & 5.5 & 3.2 \\
		\midrule[.8pt]
		\midrule[.8pt]
		Train & I-A$_1$ & D-A$_1$ & D-S$_1$ & Phone & Skirt & Toner \\
		\midrule
		\multirow{1}{*}{DM{-HYB}}
		& 86 & 434 & 958 & 1,418  & 2,984 & 1,473 \\
		\multirow{1}{*}{HierMatcher}
		& 37 & 139 & 309 & 3,799  & 2,809 & 1,082 \\
		\multirow{1}{*}{\hero+\kat$_{\text{ID3}}$}
		& 344 & 819 & 1,085 & 1,097  & 1,669 & 968 \\
		\midrule[.8pt]
		\midrule[.8pt]
		Test & I-A$_1$ & D-A$_1$ & D-S$_1$ & Phone & Skirt & Toner \\
		\midrule
		\multirow{1}{*}{DM{-HYB}}
		& 2.4 & 31.7 & 67.1 & 56.9  & 229.6 & 113.9 \\
		\multirow{1}{*}{HierMatcher}
		& 2.0 & 25.1 & 50.1 & 113.0  & 181.1 & 74.4 \\
		\multirow{1}{*}{\hero+\kat$_{\text{ID3}}$}
		& \textbf{0.4} & \textbf{1.0} & \textbf{1.4} & \textbf{2.2} & \textbf{5.4} & \textbf{3.1} \\
		\bottomrule
\end{tabular}
}
\caption{
(Epoch) Training time for one epoch \& (Train) Training time until finish \& (Test) Testing time. All the results are recorded in seconds.
}
\label{tab:efficiency}
\end{table}

\begin{table}
\centering
\scalebox{0.73}{
\setlength{\tabcolsep}{1.6pt} 
\begin{tabular}{l c c c c c c c c c }
\toprule
{Methods} & {I-A$_1$} & {D-A$_1$} & {D-S$_1$} & {I-A$_2$} & {D-A$_2$} & {D-S$_2$} & {Phone} & {Skirt} & {Toner} \\
\midrule
{ DM\footnotesize{-RNN} } & 83.1 & 98.8 & 93.5 & 67.1 & 94.8 & 89.6 & 98.2 & 91.6 & 90.9 \\
{ DM\footnotesize{-ATT} } & 83.8 & 98.8 & 93.7 & 62.2 & 94.1 & 90.4 & 95.7 & 93.2 & 91.6 \\
{ DM\footnotesize{-HYB} } & 83.5 & 98.8 & 95.0 & 64.0 & 95.9 & 92.6 & 98.7 & 94.2 & 92.0 \\
{ HierMatcher } & 79.1 & 98.5 & 94.3 & 77.1 & 96.1 & 93.0 & 96.5 & 95.4 & 94.7 \\
\midrule
{ \hero+DT } & 95.5 & 97.6 & 91.7 & 60.0 & 87.8 & 77.1 & 97.5 & 99.7 & 99.8 \\
{ \hero+\kat$_{\text{ID3}}$ } & 95.9 & 98.1 & 90.2 & 59.3 & 89.7 & 80.5 & 94.9 & 99.3 & 99.6 \\
{ \hero+\kat$_{\text{XGB}}$ } & 95.5 & 98.1 & 90.1 & 63.3 & 89.3 & 80.4 & 96.5 & 99.7 & 99.9 \\
\bottomrule
\end{tabular}
}
\caption{F$_1$ scores of all methods under sufficient resource setting(\%).}
\label{tab:full}
\end{table}

\paragraph{Sufficient Resource EM.}

Table~\ref{tab:full} shows the results with sufficient training data following 
the split method of~\citet{mudgal19deepmatcher,fu2020hierarchical}.
Our method outperforms other methods on 4 datasets, and slightly fall behind 
on 5 datasets. 
\section{Related Works}\label{sec:relate}

The way of extracting comparison features falls into two categories: 
monotonic and non-monotonic.
Monotonic features are (negatively) proportional similarities between 
attribute values. 
They can be 
calculated by symbolic rules, such as Jaccard similarity, Levenshtein similarity~\citep{fan2009reasoning,wang2011entity,conda16magellan,singh2017generating}, or learned from differentiable comparison operations, such as subtracting, point-wise multiplication
~\citep{fu2019end,ebraheem2018distributed,fu2019end}.
Non-monotonic features are hidden 
representations of end-to-end neural networks, such as {\em Softmax} or {\em Sigmoid} based similarity scores~\citep{fu2020hierarchical}, attention based scores~\citep{nie2019deep}, or simply embedding based features~\citep{mudgal19deepmatcher,li2020deep}.



EM with limited resources 
has recently intrigued research interest 
~\citep{thirumuruganathan2018reuse,kasai2019low}.
Existing explorations seek solution 
from leveraging external data to improving annotation efficiency.
External data can be aggregated via transfer learning~\citep{zhao10autoem,thirumuruganathan2018reuse,kasai2019low,loster2021knowledge}, or via pre-training language models~\citep{li2020deep}.
For better annotations, researchers tried 
active learning~\citep{kasai2019low,nafa2020active,sarawagi2002interactive,arasu2010active}, or crowd sourcing techniques~\citep{wang2012crowder,gokhale2014hands}.


The interpretability of neural models will contribute to the trust and the safety. It has become one of the central issues in machine learning.
\citet{chen2020towards} examines interpretability in EM risk analysis.
There are also attempts to explain from the perspective of attention coefficients~\citep{mudgal19deepmatcher,nie2019deep}.

\section{Conclusion}\label{sec:conclusion}

We present a decoupled framework for interpretable entity matching.
It is robust to both noisy heterogeneous input and the scale of training resources.
Experiments show that our method can be converted to interpretable rules, 
which can be inspect by domain experts and make EM process more reliable.

In the future, it is intriguing to explore more efficient ways to explore unlabeled data, such as levering connections among entities, or combine with pre-trained language models.
It is also valuable to explore how to use our heterogeneous information fusion module to boost other EM methods, such as injecting \hero~representation as supplementary information into end-to-end models.



\section*{Acknowledgments}
This work is supported by Science and Technology Innovation 2030 - New Generation of Artificial Intelligence Project (2020AAA0106501), the NSFC Key Project (U1736204), the NSFC Youth Project (62006136), the Federal Ministry of Education and Research of Germany as part of the competence center for machine learning ML2R (01IS18038C), and the grant from Alibaba Inc.


\section*{Ethical Considerations}
\paragraph{Intended Use.}
The reported technique is intended for reliable entity matching in large scale E-commercial products, where attribute values are mostly heterogeneous descriptive sentences.
The `low resource' feature is intended to avoid heavy labor force.
The `interpretability' is intended to risk control in entity matching.

\paragraph{Misuse Potential.}
As matching/alignment technique, our method may be misused in matching private information.

\paragraph{Failure Modes.}
Our method provides a promising way to have domain experts check the generated rules, thus reducing the failure risk.

\paragraph{Energy and Carbon Costs.}
The efficiency test in Section~\ref{sec:discuss} shows that our method costs less computations and is more energy saving than existing methods.

\bibliographystyle{acl_natbib}
\bibliography{1-ref}


\appendix
\phantom{a} \\ \phantom{a} \\ \phantom{a} \\ \phantom{a} \\

\begin{table*}[ht]
\centering
\small
\scalebox{1.0}{
\setlength{\tabcolsep}{7.0pt} 
\begin{tabular}{l ccc ccc ccc }
\toprule
\multirow{2}{*}{Methods} & \multicolumn{3}{c}{I-A$_1$} & \multicolumn{3}{c}{D-A$_1$} & \multicolumn{3}{c}{D-S$_1$} \\
\cmidrule(r){2-4}\cmidrule(lr){5-7}\cmidrule(lr){8-10}
 & P & R & F$_1$ & P & R & F$_1$ & P & R & F$_1$ \\
\midrule
{ DM\footnotesize{-RNN} } & 69.1 & 60.9 & 63.6 & 81.7 & 90.3 & 85.4 & 69.9 & 80.9 & 74.8 \\
{ DM\footnotesize{-ATT} } & 54.2 & 58.4 & 55.8 & 75.3 & 91.2 & 82.5 & 75.0 & 83.5 & 79.0 \\
{ DM\footnotesize{-HYB} } & 58.4 & 64.1 & 60.9 & 84.3 & 89.2 & 86.6 & 74.3 & 82.4 & 78.0 \\
{ HierMatcher } & 64.1 & 61.8 & 61.9 & 41.6 & 38.9 & 37.5 & 72.1 & 67.2 & 68.2 \\
{ Magellan } & 92.3 & 92.7 & 92.3 & 95.4 & 92.2 & 93.7 & 80.7 & \textbf{90.2} & 85.1 \\
\midrule
{ \hero+LN } & 84.1 & 73.0 & 77.9 & 15.0 & 97.1 & 21.0 & \textbf{96.1} & 44.3 & 54.7 \\
{ \hero+LR } & 79.9 & 89.1 & 84.2 & 86.7 & 95.7 & 87.1 & 85.2 & 84.2 & 84.6 \\
\midrule
{ \hero+DT } & \textbf{97.1} & \textbf{94.9} & \textbf{96.0} & \textbf{95.9} & 97.0 & 96.4 & 90.0 & 85.1 & 87.5 \\
{ \hero+\kat$_{\text{ID3}}$ } & \textbf{97.1} & 94.7 & 95.8 & 95.8 & \textbf{97.4} & \textbf{96.6} & 87.8 & 88.7 & \textbf{88.2} \\
{ \hero+\kat$_{\text{XGB}}$ } & 87.7 & 94.0 & 90.6 & 91.1 & 95.7 & 93.3 & 88.4 & 87.4 & 87.9 \\
\midrule[1pt]
\midrule[1pt]
\multirow{2}{*}{Methods} & \multicolumn{3}{c}{I-A$_2$} & \multicolumn{3}{c}{D-A$_2$} & \multicolumn{3}{c}{D-S$_2$} \\
\cmidrule(r){2-4}\cmidrule(lr){5-7}\cmidrule(lr){8-10}
 & P & R & F$_1$ & P & R & F$_1$ & P & R & F$_1$ \\
\midrule
{ DM\footnotesize{-RNN} } & 43.3 & 42.4 & 42.3 & 39.1 & 55.5 & 45.7 & 31.9 & 50.7 & 39.0 \\
{ DM\footnotesize{-ATT} } & 46.4 & 50.4 & 46.5 & 42.5 & 48.3 & 45.2 & 55.5 & 60.4 & 57.8 \\
{ DM\footnotesize{-HYB} } & 51.1 & 54.5 & 49.5 & 48.8 & 44.6 & 46.2 & 57.3 & 65.1 & 60.4 \\
{ HierMatcher } & 41.2 & 43.9 & 37.8 & 48.5 & 27.8 & 32.6 & 50.4 & 44.1 & 45.8 \\
{ Magellan } & 51.8 & 49.4 & 50.6 & 58.5 & 74.8 & 65.6 & 72.6 & 69.7 & 71.1 \\
\midrule
{ \hero+LN } & 54.1 & 34.0 & 41.6 & - & - & - & 73.1 & 84.7 & 78.5 \\
{ \hero+LR } & 49.5 & 44.5 & 46.5 & - & - & - & 62.1 & 75.7 & 68.1 \\
\midrule
{ \hero+DT } & \textbf{55.6} & \textbf{54.5} & \textbf{54.9} & \textbf{75.4} & 85.5 & 80.1 & 77.8 & 70.9 & 74.2 \\
{ \hero+\kat$_{\text{ID3}}$ } & 50.6 & 53.4 & 51.6 & 73.6 & 85.4 & 79.0 & 81.9 & \textbf{77.2} & \textbf{79.5} \\
{ \hero+\kat$_{\text{XGB}}$ } & 35.9 & 51.0 & 41.5 & \textbf{75.4} & \textbf{86.1} & \textbf{80.3} & \textbf{82.1} & 77.1 & \textbf{79.5} \\
\midrule[1pt]
\midrule[1pt]
\multirow{2}{*}{Methods} & \multicolumn{3}{c}{Phone} & \multicolumn{3}{c}{Skirt} & \multicolumn{3}{c}{Toner} \\
\cmidrule(r){2-4}\cmidrule(lr){5-7}\cmidrule(lr){8-10}
 & P & R & F$_1$ & P & R & F$_1$ & P & R & F$_1$ \\
\midrule
{ DM\footnotesize{-RNN} } & 88.1 & 92.1 & 90.0 & 62.3 & 73.8 & 67.6 & 60.3 & 80.8 & 68.6 \\
{ DM\footnotesize{-ATT} } & 77.1 & 83.8 & 80.3 & 44.5 & 70.1 & 54.4 & 40.6 & 62.2 & 48.8 \\
{ DM\footnotesize{-HYB} } & 93.9 & 90.1 & 91.9 & 55.6 & 76.1 & 64.2 & 55.0 & 87.3 & 67.4 \\
{ HierMatcher } & 83.6 & 89.2 & 86.2 & 51.7 & 77.0 & 61.7 & 46.7 & 67.9 & 55.2 \\
{ Magellan } & 95.1 & 92.1 & 93.6 & 96.1 & \textbf{97.2} & 96.6 & 96.7 & \textbf{97.6} & \textbf{97.2} \\
\midrule
{ \hero+LN } & 80.5 & 65.5 & 72.2 & 93.8 & 51.5 & 62.8 & 88.4 & 83.8 & 86.0 \\
{ \hero+LR } & \textbf{97.3} & 80.0 & 87.5 & \textbf{99.9} & 26.4 & 41.7 & 62.6 & 89.8 & 62.0 \\
\midrule
{ \hero+DT } & 93.0 & \textbf{97.0} & \textbf{94.9} & 96.7 & 96.7 & 96.7 & \textbf{97.6} & 96.7 & \textbf{97.2} \\
{ \hero+\kat$_{\text{ID3}}$ } & 92.2 & 96.9 & 94.5 & 96.9 & 96.6 & \textbf{96.7} & \textbf{97.6} & 96.7 & \textbf{97.2} \\
{ \hero+\kat$_{\text{XGB}}$ } & 92.6 & 96.1 & 94.4 & 99.0 & 93.5 & 96.2 & \textbf{97.6} & 96.8 & \textbf{97.2} \\
\bottomrule
\end{tabular}
}
\caption{Experimental results under low-resource setting with precision, recall, and F$_1$ measure (\%).
Dash (-) indicates these methods fail to converge on the datasets.
}
\label{tab:app:more}
\end{table*}

\section{More Experimental Results}

Table~\ref{tab:app:more} in the main text only shows the F$_1$ measure of the all the methods. Here, we supplement the experimental results with \textit{precision} (P = $\frac{\text{TP}}{\text{TP}+\text{FP}}$), \textit{recall} (R = $\frac{\text{TP}}{\text{TP}+\text{FN}}$) on the 9 datasets for more comprehensive analysis.
Experimental results are listed in Table~\ref{tab:app:more}.
Our methods achieve the highest precision and recall on most of the datasets.

\section{Reproducibility Details}

Each epoch of \shero training is evenly divided into 3 batches.
The \textit{Title} attribute values were padded to $l=64$, and the other attribute values are all padded to $l=32$. 
We modify the padding size on large datasets, so that our the experiments can be conducted on a single GPU.
Chinese datasets are embedded with Tencent Embedding \cite{song2018directional} and English datasets use fastText embeddings \cite{bojanowski2017enriching}.
Multi-head mechanism is used in the attention module.
The embedding size $d_e$ for Chinese is 300, and for English is 200.
AGG converts embedding into $d_a$ dimensional vectors, where $d_a = 100$.
PROP further outputs with a 2-layer MLP with dimension size $d = 64$.
The query vector and the key vector in the attention layer of PROP are 16 dimensional vectors.
During training, attribute values are masked at a probability $p = 0.4$.
The  Adam optimizer \cite{kingma2014adam} is used for \shero.
Training rate and L$_2$ weight decay are 0.01 and $10^{-5}$.

\kat$_\text{XGB}$ is implemented using {\tt xgboost 0.9} with objective function \textit{binary: logistic}.
\kat$_\text{ID3}$ is implemented using  {\tt scikit-learn 0.24}.
\shero is implemented with {\tt PyTorch 1.4.0} in {\tt Python 3.7.6}.
The comparison feature metrics in Table~\ref{tab:app:metrics} are implemented with {\tt py-entitymatching 0.4.0}.
We also use {\tt Numpy 1.19.2} for matrix calculation.
All the experiments are evaluated on a single NVIDIA 3090 GPU with 24GiB GRAM.

\end{document}